\documentclass[letterpaper, 10 pt, conference]{Classes/ieeeconf}  
\IEEEoverridecommandlockouts 
\usepackage{Packages/decar-common}
\usepackage{Packages/decar-dynamics}
\usepackage[backend=bibtex,bibstyle=ieee,citestyle=numeric-comp,doi=false,isbn=false]{biblatex}
\usepackage{graphicx}
\usepackage{lipsum}
\usepackage{derivative}
\usepackage{soul}
\usepackage[colorlinks, citecolor=black, linkcolor=black, linktocpage=true, hidelinks]{hyperref}
\usepackage{booktabs}
\usepackage{tabularx}
\usepackage{threeparttable}
\usepackage{float}
\usepackage{dblfloatfix}
\usepackage{placeins}
\usepackage{xcolor}
\usepackage{orcidlink}

\bibliography{refs}
\graphicspath{{figs/}}

\title{\LARGE \bf
    Observability-Informed Optimal Sensor Placement for Soft Robots
}

\author{
    Samuel Smocot$^{1}$, James R. Forbes$^{1,2}$, Audrey Sedal$^{1,2}$
\thanks{This work was supported by the Fonds de Recherche du Québec - Nature et Technologies and the NSERC Discovery Grant program.}
\thanks{$^{1}$ Dept. of Mechanical Engineering, McGill University, Montreal, Canada.}
\thanks{$^{2}$ Mila -- Québec AI Institute, Montreal, Canada.}
\thanks{Contact: {\tt\small samuel.smocot@mail.mcgill.ca}}
\thanks{Source code: \href{https://github.com/macrobotics-lab/obsv-sens-plcmt}{https://github.com/macrobotics-lab/obsv-sens-plcmt}}
}

\begin{document}
%
%
%
%
%
%
%
\def \myJournal {2026 IEEE 9th International Conference on Soft Robotics (RoboSoft)}
\def \myDoi {10.1109/RoboSoft67810.2026.11522828}
\def \myPaperSiteName {IEEE Xplore}
\def \myPaperSiteLink {https://ieeexplore.ieee.org/document/11522828}
\def \myYear {2026}
\def \myPaperCitation{S. Smocot, J. R. Forbes and A. Sedal, "Observability-Informed Optimal Sensor Placement for Soft Robots," 2026 IEEE 9th International Conference on Soft Robotics (RoboSoft), Kanazawa, Japan, 2026, pp. 430-437, doi: 10.1109/RoboSoft67810.2026.11522828.}


\begin{figure*}[t]

\thispagestyle{empty}
\begin{center}
\begin{minipage}{6in}
\centering
This paper has been accepted for publication in \emph{\myJournal}. 
\vspace{1em}

This is the author's version of an article that has, or will be, published in this journal or conference. Changes were, or will be, made to this version by the publisher prior to publication.
\vspace{2em}

\begin{tabular}{rl}
DOI: & \myDoi\\
\myPaperSiteName: & \texttt{\myPaperSiteLink}
\end{tabular}

\vspace{2em}
Please cite this paper as:

\myPaperCitation

\vspace{15cm}
\copyright \myYear \hspace{4pt}IEEE. Personal use of this material is permitted. Permission from IEEE must be obtained for all other uses, in any current or future media, including reprinting/republishing this material for advertising or promotional purposes, creating new collective works, for resale or redistribution to servers or lists, or reuse of any copyrighted component of this work in other works.

\end{minipage}
\end{center}
\end{figure*}
\newpage
\clearpage
\pagenumbering{arabic} 
    \maketitle
    \thispagestyle{empty}
    \pagestyle{empty}
    
    \begin{abstract}
    This paper presents the application and experimental evaluation of a systematic method for optimal sensor placement in soft robots.
    Existing methods either lack generalizability across different soft robot morphologies or do not account for system dynamics.
    The applied method uses convex optimization to find the optimal sensor configuration that maximizes an observability Gramian-based metric. 
    The framework is experimentally evaluated using position and strain measurements on a soft continuum arm.
    Kalman filter state estimates using optimal sensor placements yield lower reconstruction error than a baseline across all sinusoidal input trials, with improvements on the order of millimeters.
    This case study shows that linear control theory tools can guide optimal sensor placement in soft robots, suggesting an interpretable approach to sensor placement that may extend to other morphologies.
\end{abstract}
    \vspace{-4pt}
\section{Introduction}
\label{sec:introduction}

Soft robot sensing is an open challenge due to soft robots' continuous deformability and high number of passive degrees of freedom (DoF), which also make open-loop control difficult for precise tasks~\cite{wangPerceptiveSoftRobots2018}. 
State estimation and closed-loop control necessary for complex tasks require efficient sensor integration~\cite{linRecentAdvancesPerceptive2023}, yet unlike rigid robots, selecting, integrating and building perception frameworks for sensors on soft robots is difficult.
Practical challenges related to electronics, sensor size, and limited spaces prevent dense sensor integration~\cite{wangPerceptiveSoftRobots2018}. 
Certain placements may compromise the robot's softness while hindering its ability to bend, stretch, and move freely~\cite{hegdeSensingSoftRobotics2023}.
Advances in sensor technologies that miniaturize, add mechanical compliance, or capture multiple DoF render the number of possible placement locations and integration overwhelming.
Exhaustively searching all possible sensor configurations can then be computationally intractable.
This NP-hard combinatorial problem grows with the number of potential sensor locations~\cite{bruntonSparseSensorPlacement2016}.
A systematic approach to optimal sensor placement for soft robots is therefore needed.

Some existing approaches to sensor placement in soft robots optimize for specific morphologies. One approach determines the placement of fiber Bragg grating sensors by minimizing a continuum robot's reconstruction error~\cite{kimOptimizingCurvatureSensor2014}. Another method minimizes estimation variance at selected points along a robot's arc length~\cite{mahoneyInseparableNatureSensor2016}. An experimental framework for soft robotic hands placed sensors via prediction accuracy of manipulation success and object pose~\cite{liSoftTouchSensorPlacementFramework2022}. Nonlinear observers show the possibility of full-state estimation of a continuum robot using only tip velocity~\cite{zheng2024full}.

Several methods are not tied to a particular soft robot morphology. 
One learns the mapping from redundant sensor data to actuator deformation, then reduces the sensor layout to an optimal configuration \cite{wallMethodSensorizingSoft2017}. Similarly, another incrementally selects sensors from a large set based on the reconstruction performance gain~\cite{tapiaMakeSenseAutomatedSensor2020}.
Another approach selects a fixed number of sensors from a candidate set by maximizing a task objective using Thompson sampling~\cite{kimOptimalSensorPlacement2024}.
While not tied to specific robot morphologies, these approaches neglect the robot's dynamics. 
A method addressing both generalizability and dynamics uses a neural network (NN) to learn optimal sensor locations alongside a task-specific model~\cite{spielbergCoLearningTaskSensor2021}. However, it only supports strain sensors and relies on the material point method, which is difficult to experimentally validate on real-world robots due to computational limitations~\cite{navezModelingEmbeddedControl2025}.
The NN approach also requires extensive data collection, which can be time-consuming and cumbersome.

\begin{figure}[!t]
    \centering
    \includegraphics[width=\linewidth]{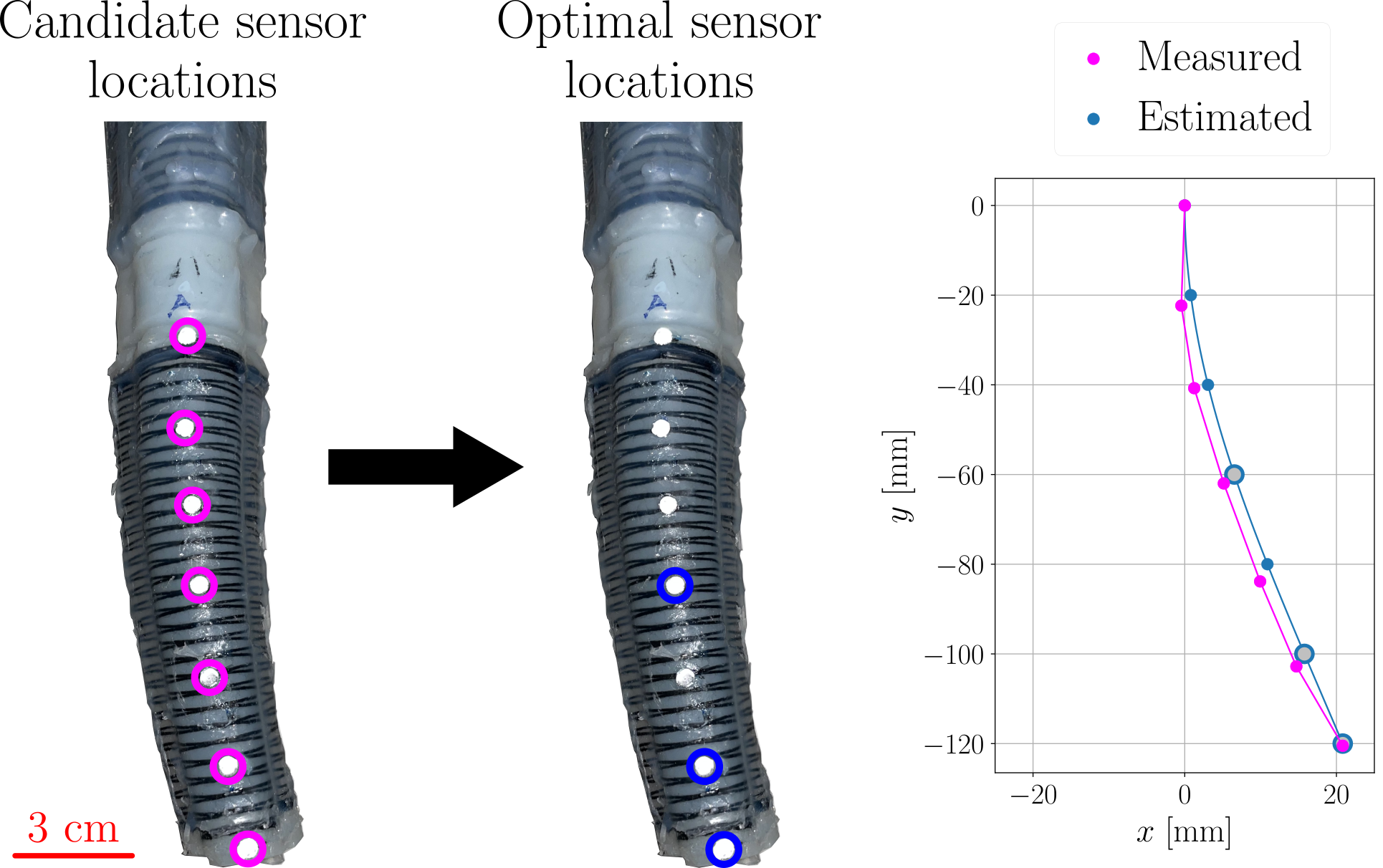}
    \vspace{-10pt}
    \caption{Overview of the sensor placement framework: starting from candidate sensor locations, the method selects an optimal configuration to maximize observability. The shape reconstruction comparison illustrates the quality of the selected placement.}
    \label{fig:summary}
    \vspace{-20pt}
\end{figure}

In contrast, observability-based methods incorporating system dynamics for sensor placement have shown success outside of soft robotics. 
A convex relaxation of the observability Gramian-based optimization problem for sensor selection was proposed in~\cite{summersConvexRelaxationsGramian2016}. 
An empirical observability Gramian-based optimization was applied to an aircraft model~\cite{hodzicSimulationBasedObservabilityAnalysis2021}, and later extended to continuous systems via a cantilever beam~\cite{braceSensorPlacementCantilever2022}. 
While these methods provide a general framework for sensor placement, their application to soft robots has not been evaluated.

This work contributes the first implementation and experimental evaluation of the Gramian-based optimization problem formulation in~\cite{summersConvexRelaxationsGramian2016} on a physical soft robot, using a continuum robot as a representative example. Fig.~\ref{fig:summary} illustrates the framework from candidate sensor locations to the resulting placement. 
Placements obtained from this dynamics-aware observability-based method yield lower or comparable RMSE to a baseline of evenly spaced sensors when estimating the robot's shape.
A background on observability and the sensor placement optimization problem are described in Section~\ref{sec:sensor_placement_methodology}. Section~\ref{sec:application} details the experimental implementation on a continuum arm, followed by the results for sensor placement and estimation in Section~\ref{sec:results}. Section~\ref{sec:discussion} discusses findings and limitations, and Section~\ref{sec:conclusion} concludes the paper.

    \section{Sensor Placement Methodology}
\label{sec:sensor_placement_methodology}

This section describes the sensor placement framework from~\cite{summersConvexRelaxationsGramian2016}, which solves an optimization problem to determine optimal sensor locations by maximizing system observability. Additional details are included to handle multiple sensing modalities, which the original work did not cover.

\subsection{Observability Metrics}
\label{sec:observability_tools}

The state-space form of a linear time-invariant system with no feedthrough is given by
\begin{equation}
    \begin{aligned}
        \dot{\mbf{x}}(t) &= \mbf{Ax}(t) + \mbf{Bu}(t), \\
        \mbf{y}(t) &= \mbf{Cx}(t),
    \end{aligned}
    \label{eq:state-space}
\end{equation}
where $\mbf{x}(t)$ is the state vector, $\mbf{u}(t)$ is the input, and $\mbf{y}(t)$ is the output. The $n$-dimensional system described in \eqref{eq:state-space} is observable if the system's state can be uniquely determined from the observed outputs~\cite{willemsControllabilityObservabilityPole1971}. This condition is satisfied if and only if the observability matrix, given by
\begin{equation}
    \mbf{Q}_{\mathrm{o}} = \begin{bmatrix}
        \mbf{C} \\ \mbf{CA} \\ \vdots \\ \mbf{CA}^{n-1}
    \end{bmatrix},
    \label{eq:observability_matrix}
\end{equation}
is full rank~\cite{willemsControllabilityObservabilityPole1971}. The rank of the observability matrix only indicates whether a system is observable. The degree of observability can be quantified by analyzing the observability Gramian $\mbf{W}_{\mathrm{o}}$, which is the solution to the Lyapunov equation
\begin{equation}
    \mbf{A}^{\trans} \mbf{W}_{\mathrm{o}} + \mbf{W}_{\mathrm{o}} \mbf{A} + \mbf{C}^{\trans} \mbf{C} = \mbf{0}.
    \label{eq:Lyapunov}
\end{equation}

The observability Gramian $\mbf{W}_{\mathrm{o}}$ is a measure of how much energy is transferred from the states to the output, and therefore measures ``how observable'' a given state is \cite{dullerudCourseRobustControl2000}. A system's degree of observability is maximized using certain metrics on $\mbf{W}_{\mathrm{o}}$. Commonly used metrics include~\cite{hodzicSimulationBasedObservabilityAnalysis2021, summersConvexRelaxationsGramian2016}:
\begin{itemize}
    \item $\lambda_{\min}(\mbf{W}_{\mathrm{o}})$, which aims to maximize the observability of the least observable mode of the system, thereby indicating how near the system is to being unobservable~\cite{hodzicSimulationBasedObservabilityAnalysis2021}.
    \item $\trace(\mbf{W}_{\mathrm{o}})$, which maximizes the energy transfer from the perturbations to the measurements. This metric does not guarantee nonzero eigenvalues~\cite{hodzicSimulationBasedObservabilityAnalysis2021}.
    \item $\log(\det(\mbf{W}_{\mathrm{o}}))$, which maximizes the product of the eigenvalues of $\mbf{W}_{\mathrm{o}}$ and in turn the overall observability of the system~\cite{hodzicSimulationBasedObservabilityAnalysis2021}. 
\end{itemize}

\vspace{-6pt}
\subsection{Optimization Problem}
\vspace{-2pt}
\label{sec:optimization_problem}
As seen in \eqref{eq:Lyapunov}, the system's observability depends on the dynamics matrix $\mbf{A}$ and the measurement matrix $\mbf{C}$. The optimal sensor placement involves designing $\mbf{C}$ so that the pair $(\mbf{A}, \mbf{C})$ results in a $\mbf{W}_{\mathrm{o}}$ that maximizes observability.

Here, an initial $\mbf{C}$ is constructed, where each row corresponds to a candidate sensor location. Optimally placing $p$ active sensors involves selecting an optimal subset of $p$ rows from $\mbf{C}$, defined by a binary selector vector $\mbs{\alpha} \in \{0, 1\}^{r}$, where $r$ is the number of candidate locations. The goal is to find $\mbs{\alpha}$ such that the resulting matrix $\mbf{C}^{*} = \textrm{diag}(\mbs{\alpha}) \mbf{C}$ maximizes a chosen metric on the observability Gramian $\mathcal{J}(\mbf{W}_{\mathrm{o}})$. The optimization problem is then formulated as
\vspace{-4pt}
\begin{subequations}
    \begin{align}
        \max_{\mbs{\alpha}} &\quad \mathcal{J}(\mbf{W}_{\mathrm{o}}) \\
        \text{such that} &\quad \mbf{W}_{\mathrm{o}} = \mbf{W}_{\mathrm{o}}^{\trans} \succ 0, \\
        &\quad \mbf{A}^{\trans}\mbf{W}_{\mathrm{o}} + \mbf{W}_{\mathrm{o}}\mbf{A} + (\mathrm{diag}(\mbs{\alpha})\mbf{C})^{\trans} \mathrm{diag}(\mbs{\alpha}) \mbf{C} = \mbf{0}, \label{eq:lyapunov_nonconvex}\\
        &\quad \mbs{\alpha} \in \{0, 1\}^{r}, \label{eq:set_constraint}\\
        &\quad \begin{bmatrix}
            1 & 1 & \cdots & 1
        \end{bmatrix} \mbs{\alpha} - p = 0.
    \end{align}
    \label{eq:opt_prob_nonconvex}
\end{subequations}
\vspace{-1pt}
The optimization problem~\eqref{eq:opt_prob_nonconvex} is nonconvex due to the constraints~\eqref{eq:lyapunov_nonconvex} and~\eqref{eq:set_constraint}. Following the approach in~\cite{summersConvexRelaxationsGramian2016} for controllability Gramian optimization, the term $(\mathrm{diag}(\mbs{\alpha})\mbf{C})^{\trans} \mathrm{diag}(\mbs{\alpha}) \mbf{C}$ is rewritten as $\sum_{i=1}^{r}\alpha_i\mbf{c}_{i}^{\trans}\mbf{c}_{i}$ where $\mbf{c}_{i}$ is the $i$th row of $\mbf{C}$. This equivalence holds when all elements of $\mbs{\alpha}$ are either 0 or 1. Also, the set constraint on $\mbs{\alpha}$ is relaxed to obtain a convex optimization problem. The resulting convex optimization is given by
\vspace{-4pt}
\begin{subequations}
    \begin{align}
        \max_{\mbs{\alpha}} &\quad \mathcal{J}(\mbf{W}_{\mathrm{o}}) \\
        \text{such that} &\quad \mbf{W}_{\mathrm{o}} = \mbf{W}_{\mathrm{o}}^{\trans} \succ 0, \\
        &\quad \mbf{A}^{\trans}\mbf{W}_{\mathrm{o}} + \mbf{W}_{\mathrm{o}}\mbf{A} + \sum_{i=1}^{r}\alpha_i\mbf{c}_{i}^{\trans}\mbf{c}_{i} = \mbf{0}, \\
        &\quad 0 \leq \alpha_{i} \leq 1, \\
        &\quad \begin{bmatrix}
            1 & 1 & \cdots & 1
        \end{bmatrix} \mbs{\alpha} - p = 0.
    \end{align}
    \label{eq:opt_prob_convex}
\end{subequations}
\vspace{-1pt}
When the solution to the relaxed problem satisfies the integer constraint of the initial integer programming (IP) problem, then this solution is optimal for the IP problem as well~\cite{hillierIntroductionOperationsResearch2024}. This means that if the solution $\mbs{\alpha}$ to the relaxed problem~\eqref{eq:opt_prob_convex} satisfies the set constraint \eqref{eq:set_constraint}, then it is also the solution to the original nonconvex problem.
Here, a tolerance of 0.15 is used to determine whether the solution satisfies the set constraint.
The advantage of convex optimization is that it is efficient, scales well to large problems, and guarantees a globally optimal solution~\cite{boydConvexOptimization2004}.

Additionally, the formulation in~\eqref{eq:opt_prob_convex} also allows for multiple concurrent sensing modalities. Let $\mbf{C}_{1}$ and $\mbf{C}_{2}$ correspond to 2 different sensing modalities, and $\mbs{\alpha}_{1}$ and $\mbs{\alpha}_{2}$ be their respective selector vectors. The initial matrix with all candidate locations is then $\begin{bmatrix}
    \mbf{C}_{1}^{\trans} & \mbf{C}_{2}^{\trans}
\end{bmatrix}^{\trans}$, and the optimization problem formulation for two sensing modalities is then
\begin{subequations}
    \begin{align}
        \max_{\mbs{\alpha}_{1}, \mbs{\alpha}_{2}} &\quad \mathcal{J}(\mbf{W}_{\mathrm{o}}) \\
        \text{such that} &\quad \mbf{W}_{\mathrm{o}} = \mbf{W}_{\mathrm{o}}^{\trans} \succ 0, \\
        &\quad \mbf{A}^{\trans}\mbf{W}_{\mathrm{o}} + \mbf{W}_{\mathrm{o}}\mbf{A} + \sum_{i=1}^{r_{1}}\alpha_{1,i}\mbf{c}_{1,i}^{\trans}\mbf{c}_{1,i} \nonumber\\
        & \hspace{2.5cm} + \sum_{i=1}^{r_{2}}\alpha_{2,i}\mbf{c}_{2,i}^{\trans}\mbf{c}_{2,i} = \mbf{0}, \\
        &\quad 0 \leq \alpha_{1,i} \leq 1, \quad 0 \leq \alpha_{2,i} \leq 1, \\
        &\quad \begin{bmatrix}
            1 & 1 & \cdots & 1
        \end{bmatrix} \mbs{\alpha}_{1} - p_{1} = 0, \\
        &\quad \begin{bmatrix}
            1 & 1 & \cdots & 1
        \end{bmatrix} \mbs{\alpha}_{2} - p_{2} = 0.
    \end{align}
    \label{eq:opt_prob_convex_2_types}
\end{subequations}
\vspace{-15pt}

\begin{samepage}
    The optimal sensor placement approach is:
    \begin{enumerate}
        \item Model the robot dynamics in state-space form.
        \item Collect candidate sensor locations in initial $\mbf{C}$.
        \item Solve optimization problem~\eqref{eq:opt_prob_convex}.
        \item Form optimal measurement matrix $\mbf{C}$ with the rows selected by the optimization problem solution.
    \end{enumerate}
\end{samepage}

For a nonlinear system, the same approach can be applied by first linearizing the system. This can be done either at a single operating point where the robot typically operates, or at multiple points along its trajectory. In the latter case, the optimization problem can be formulated to maximize observability across all linearization points.

    \section{Application to a Continuum Robot Arm}
\label{sec:application}

While the sensor placement framework is general to any system with a state-space representation, it was implemented and evaluated on a continuum robot arm, specifically a replica of the Soft continuum Proprioceptive Arm~\cite{toshimitsuSoPrAFabricationDynamical2021} without use of its proprioceptive backbone. The robot, shown in Fig.~\ref{fig:summary}, is pneumatically actuated and has two segments, each with three silicone-cast chambers. The fibers wound around each chamber induce bending towards the opposing two chambers when pressurized. This robot was selected due to its ease of fabrication~\cite{toshimitsuSoPrAFabricationDynamical2021} and because continuum arms are one of the most prevalent designs in soft robotics~\cite{russoContinuumRobotOverview2023}.

\subsection{Modelling}
\label{sec:continuum_robot_arm_modeling}

One segment of the robot was modelled as a cantilever Euler-Bernoulli beam of length $L$. The work in~\cite{braceSensorPlacementCantilever2022} also applied observability Gramian-based sensor placement to a cantilever beam, but with an empirical Gramian and did not account for damping. It considered only strain sensors, whereas both position and strain measurements are considered here. Additionally, the resulting placements were not experimentally evaluated on a physical system. A benefit of modelling the segment as an Euler-Bernoulli beam model is that it is inherently linear, allowing direct use in the optimization problem~\eqref{eq:opt_prob_convex} thus providing a useful approximation for sensor placement. More common nonlinear models for continuum arms, such as the Cosserat rod model, would require linearization. Since this pneumatically actuated arm does not undergo torsion or shear, this simpler beam model is a reasonable choice.

The equation of motion for the forced lateral vibration of a uniform beam is given by
\begin{equation}
    \varrho A \pdv[order=2]{w(x,t)}{t} + EI \pdv[order=4]{w(x,t)}{x}  = f(x,t),
    \label{eq:EoM_uniform_beam}
\end{equation}
where $\varrho$ is the mass density, $A$ is the cross-sectional area, $E$ is the Young's modulus, $I$ is the moment of inertia of the beam cross section, and $f(x,t)$ is the external force per unit length of the beam~\cite{raoMechanicalVibrations2017}. Using the Rayleigh-Ritz method, the deflection is approximated as a linear combination of $N$ basis functions $\phi_{i}(x)$ and time-dependent coefficients $q_{i}(t)$, that is, $w(x,t) \approx \sum_{i=1}^{N} \phi_{i}(x) q_{i}(t)$. The basis functions are taken to be the mode shapes of a fixed-free beam.

The kinetic and potential energy of the beam, expressed in terms of modal coordinates, are given by
\begin{align}
    T(\dot{\mbf{q}}(t)) &= \frac{1}{2} \dot{\mbf{q}}^{\trans}(t) \underbrace{\varrho \int_{0}^{L} \mbs{\Phi}^{\trans}(x) \mbs{\Phi}(x) \dee x}_{\mbf{M}} \, \dot{\mbf{q}}(t), \\
    V(\mbf{q}(t)) &= \frac{1}{2} \mbf{q}^{\trans} \underbrace{EI \int_{0}^{L} \mbs{\Psi}^{\trans}(x) \mbs{\Psi}(x) \dee x}_{\mbf{K}} \, \mbf{q}(t),
\end{align}
where $\mbf{q}(t)$ contains the coefficients $q_{i}(t)$, $\mbs{\Phi}(x)$ is the row vector of the mode shapes $\phi_{i}(x)$, and $\mbs{\Psi}(x)$ is the row vector of their second spatial derivatives. The Euler-Lagrange equation is then applied to obtain the equation of motion for the unforced, undamped system, yielding
\begin{equation}
    \mbf{M} \ddot{\mbf{q}}(t) + \mbf{K} \mbf{q}(t) = \mbf{0}.
    \label{eq:Mqddot+Kq=0}
\end{equation}
Equation~\eqref{eq:Mqddot+Kq=0} has a harmonic solution of the form \mbox{$\mbf{q}(t)=\bar{\mbf{q}} \exp(j \omega t)$}, which when substituted back into~\eqref{eq:Mqddot+Kq=0} yields the generalized eigenvalue problem
\begin{equation}
    \mbf{K} \bar{\mbf{q}} = \omega^{2} \mbf{M} \bar{\mbf{q}}.
    \label{eq:eigenvalue_problem}
\end{equation}

For better numerical conditioning, the system is written with mass-normalized coordinates $\mbf{v}_{i}$, defined as $\mbf{v}_{i} = \bar{\mbf{q}}_{i}/\sqrt{\bar{\mbf{q}}_{i}^{\trans} \mbf{M} \bar{\mbf{q}}_{i}}$. Any state vector $\mbf{q}(t)$ can be expressed as a linear combination of the eigenvectors $\mbf{v}_{i}$, such that
\begin{equation}
    \mbf{q}(t) = \sum_{i=1}^{N} r_{i}(t) \mbf{v}_{i} = \mbf{V} \mbf{r}(t),
    \label{eq:q(t)=Vr(t)}
\end{equation}
where $\mbf{V} = \begin{bmatrix}\mbf{v}_{1} & \mbf{v}_{2} & \cdots & \mbf{v}_{N} \end{bmatrix}$ and $\mbf{r}(t)$ contains the modal coordinates $r_{i}(t)$~\cite{inmanVibrationControl2017}.

Using the principle of virtual work, the components of the generalized force vector $\mbs{F}(t)$ are
\begin{equation}
    F_{i}(t) = \int_{0}^{L} f(x,t) \phi_{i}(x) \dee x.
    \label{eq:generalized_force}
\end{equation}

By adding the damping matrix $\mbf{D}$ and the generalized force $\mbs{F}(t)$ to \eqref{eq:Mqddot+Kq=0}, substituting \eqref{eq:q(t)=Vr(t)}, and premultiplying by $\mbf{V}^{\trans}$, the system becomes
\begin{equation}
    \begin{aligned}
        \mbf{V}^{\trans} \mbf{M} \mbf{V} \ddot{\mbf{r}}(t) + \mbf{V}^{\trans} \mbf{D} \mbf{V} \dot{\mbf{r}}(t) + \mbf{V}^{\trans} \mbf{K} \mbf{V} \mbf{r}(t) &= \mbf{V}^{\trans} \mbs{F}(t), \\
        \ddot{\mbf{r}}(t) + \tilde{\mbf{D}} \dot{\mbf{r}}(t) + \mbs{\Omega}^{2} \mbf{r}(t) &= \mbf{V}^{\trans} \mbs{F}(t),
    \end{aligned}
    \label{eq:EoM}
\end{equation}
where $\mbf{V}^{\trans} \mbf{M} \mbf{V} = \mbf{1}$, $\mbf{V}^{\trans} \mbf{D} \mbf{V} = \diag\{2 \zeta_{i} \omega_{i} \} = \tilde{\mbf{D}}$, and $\mbf{V}^{\trans} \mbf{K} \mbf{V} = \mbs{\Omega}^{2}$, with $\mbs{\Omega}^{2} = \diag\{\omega_{i}^{2} \}$~\cite{inmanVibrationControl2017}.

Letting \mbox{$\mbf{z}(t) = \mbs{\Omega} \mbf{r}(t)$}, the dynamics of the system can be written in state-space form as
\begin{equation}
    \begin{bmatrix}
        \dot{\mbf{z}}(t) \\
        \ddot{\mbf{r}}(t)
    \end{bmatrix} = \begin{bmatrix}
        \mbf{0} & \mbs{\Omega} \\
        -\mbs{\Omega} & -\tilde{\mbf{D}}
    \end{bmatrix} \begin{bmatrix}
        \mbf{z}(t) \\
        \dot{\mbf{r}}(t)
    \end{bmatrix} + \begin{bmatrix}
        \mbf{0} \\
        \mbf{V}^{\trans}
    \end{bmatrix}
    \mbs{F}(t).
    \label{eq:state_space_dynamics}
\end{equation}

The position sensors serve as measurements of the displacement of the beam from its neutral axis. The measurement model at point $x_{s}$ along the beam is then given by
\begin{equation}
    \begin{aligned}
        y_{p}(t) &= w(x_{s}, t) = \sum_{i=1}^{N} \phi_{i}(x_{s}) q(t) \\
        &= \mbs{\Phi}(x_{s}) \mbf{q}(t) = \mbs{\Phi}(x_{s}) \mbf{V} \mbf{r}(t) \\
        &= \begin{bmatrix}
            \mbs{\Phi}(x_{s}) \mbf{V} \mbs{\Omega}^{-1} & \mbf{0}_{1 \times N}
        \end{bmatrix} \begin{bmatrix}
            \mbf{z}(t) \\
            \dot{\mbf{r}}(t)
        \end{bmatrix}.
    \end{aligned}
    \label{eq:position_measurement_model}
\end{equation}
The strain measurement at point $x_{s}$ is given by
\begin{equation}
    \begin{aligned}
        y_{s}(t) &= h_{s} \pdv[order=2]{w(x_{s}, t)}{x} = h_{s} \sum_{i=1}^{N} \phi_{i}''(x_{s}) q(t) \\
        &= h_{s} \begin{bmatrix}
            \mbs{\Phi}''(x_{s}) \mbf{V} \mbs{\Omega}^{-1} & \mbf{0}_{1 \times N}
        \end{bmatrix} \begin{bmatrix}
            \mbf{z}(t) \\
            \dot{\mbf{r}}(t)
        \end{bmatrix},
    \end{aligned}
    \label{eq:strain_measurement_model}
\end{equation}
where $h_{s}$ is the distance from the beam's neutral axis, taken here to be the center of the beam. The row matrices in~\eqref{eq:position_measurement_model} and~\eqref{eq:strain_measurement_model} evaluated at various points along the beam form the rows of a measurement matrix $\mbf{C}$, which together with \eqref{eq:state_space_dynamics} forms the complete state-space model of the system.

\subsection{Experimental Setup}

For simplicity, the experiments used only the second segment of the robot, which was \SI{12}{\centi\meter} long. The segment's chambers were actuated with valves on a FESTO VTEM manifold. Pressure input signals were generated in Simulink and loaded to a Speedgoat Baseline Real-Time Target Machine for control. Integrated sensors in the VTEM manifold measured the valve port pressure at \SI{100}{\hertz}.

\begin{figure}[!t]
    \centering
    \includegraphics[width=0.49\linewidth]{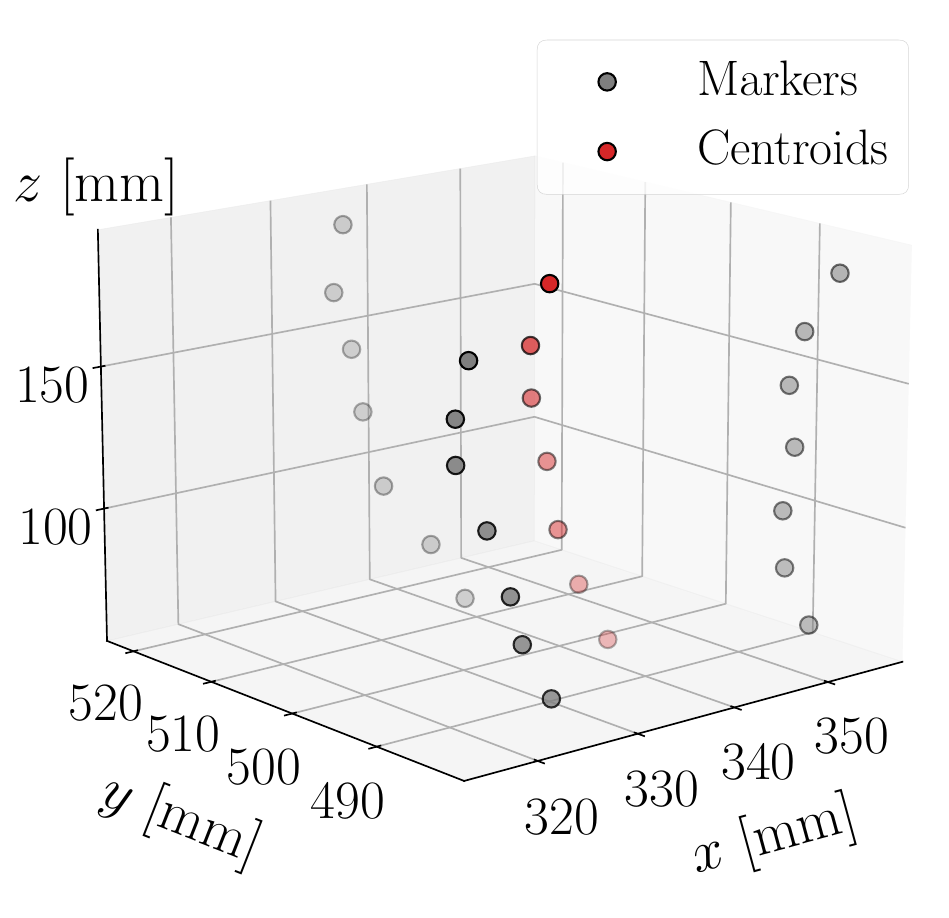}%
    \hfill
    \includegraphics[width=0.49\linewidth]{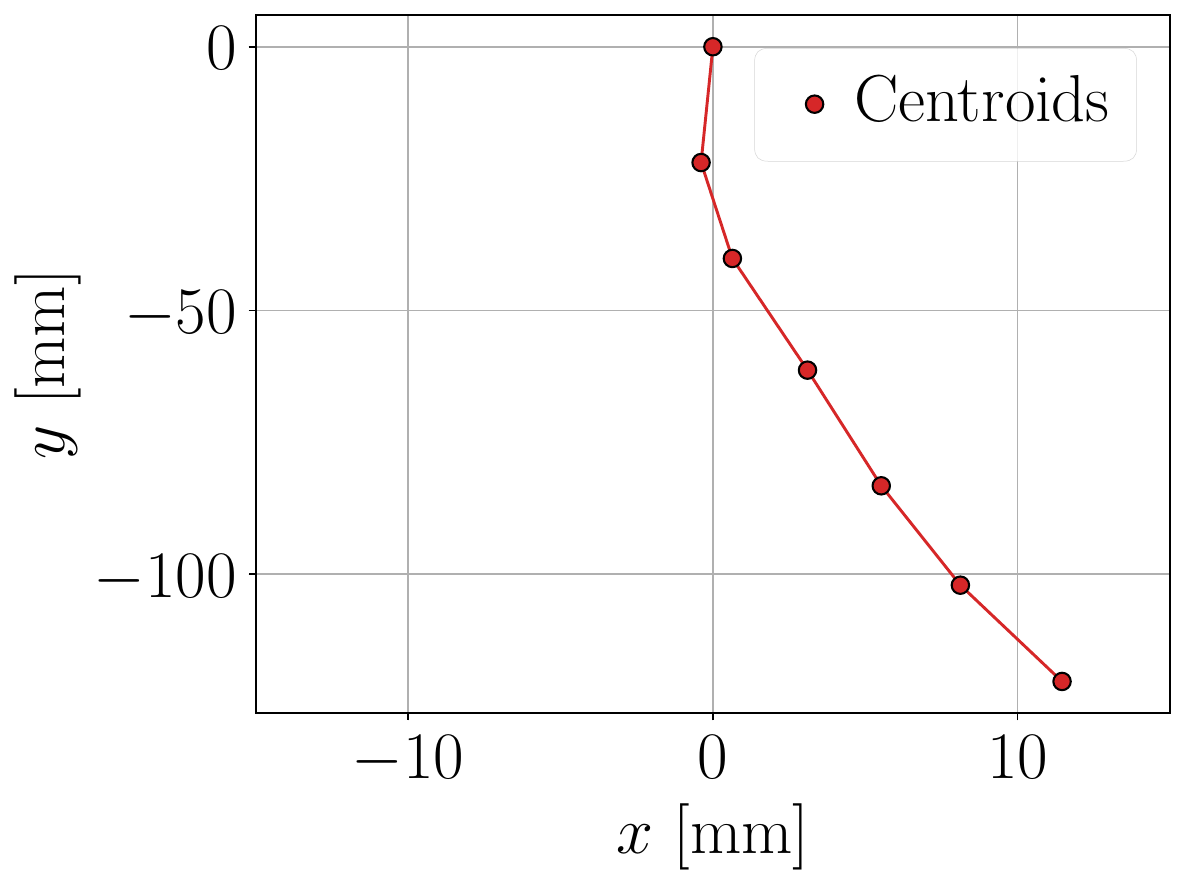}
    \caption{Marker and centroid positions on the soft robot captured by the mocap system. Left: 3D positions of markers placed along the three chambers and their computed centroids. Right: 2D projection of the centroids onto a plane-of-best-fit, enabling planar modelling of the continuum arm.}
    \label{fig:markers_and_centroids}
    \vspace{-15pt}
\end{figure}

A Vicon motion capture (mocap) system tracked the positions of seven equally spaced markers placed along each of the three chambers in the second segment. The sampling rate was \SI{100}{\hertz}. Only one chamber was actuated at a time, resulting in planar motion of the robot. Markers at corresponding heights on each of the three chambers were grouped together, and the centroid of each trio was computed. A plane-of-best-fit was then used to map the 3D centroids to 2D positions, enabling the use of the planar model described in Section~\ref{sec:continuum_robot_arm_modeling}. This process is shown in Fig.~\ref{fig:markers_and_centroids}.

\subsection{System Identification}
The natural frequencies were identified using impulse response experiments. For each trial, the actuated chamber pressure was set between \qtyrange[range-phrase={ and }]{0}{100}{\kilo\pascal} in \SI{10}{\kilo\pascal} increments, then the system was perturbed with an impulsive external input. After excitation, the tip centroid motion was tracked, and a Fourier transform was used to extract the dominant frequency components. The natural frequencies were estimated by averaging the frequency peaks across all trials. Only the first two natural frequencies were retained, as the beam model requires identification of just two parameters, $\varrho$ and $EI$, found in $\mbf{M}$ and $\mbf{K}$ respectively. As a result, the modal decomposition included only the first two mode shapes. An optimization problem was solved to identify $\varrho$~=~\SI[per-mode = symbol]{2.001}{\kilo\gram\per\meter} and $EI$~=~\SI{0.022}{\newton\meter\squared} by minimizing the difference between the experimentally identified natural frequencies and the square roots of the eigenvalues in~\eqref{eq:eigenvalue_problem}.

To identify the damping ratios in $\tilde{\mbf{D}}$, the tip centroid displacement was plotted for each impulse response trial. The logarithmic decrement $\delta$ was calculated from the first two peaks, and the first mode damping ratio was computed with $\zeta_{1} = \delta / \sqrt{4\pi^{2} + \delta^{2}}$, then averaged across trials. For simplicity, stiffness proportional damping was assumed when estimating the damping ratios from the natural frequencies, so the relation $\zeta_{i} = \beta \omega_{i} / 2$ applies~\cite{raoMechanicalVibrations2017}. The coefficient $\beta$ was determined from $\zeta_{1}$ and $\omega_{1}$, and used to compute the damping ratio of the second mode. The identified damping ratios were $\zeta_{1} = 0.181$ and $\zeta_{2} = 1.385$.

A sensitivity analysis showed that the optimal placement was robust to variations in the identified system parameters. The physical parameters $\varrho$ and $EI$, and the damping ratios $\zeta_{i}$ could vary by up to \SI{\pm 97}{\percent}, \SI{\pm 45}{\percent} and \SI{\pm 9}{\percent}, respectively, without affecting the sensor placement result and objective function value in \eqref{eq:opt_prob_convex}.

The pressure actuation was modelled as a uniformly distributed force, of the form $f(x,t) = b p_{\text{input}}$, where $p_{\text{input}}$ is the measured input pressure. The static robot position was recorded for constant pressures from \qtyrange{15}{100}{\kilo\pascal} in \SI{5}{\kilo\pascal} increments. The coefficient $b$ was identified by minimizing the root mean square error (RMSE) between the measured static position of the robot and the static position predicted by the model across 18 trials, yielding an average $b = 0.231$.

\subsection{Experimental Procedures}
\label{sec:experimental_procedures}

Step inputs from \qtyrange{0}{100}{\kilo\pascal} in \SI{5}{\kilo\pascal} increments were applied to the actuated chamber, resulting in 21 step-input trials. Additionally, sinusoidal inputs of the form $p_{\text{set}} = A (\sin(\omega t) + 1)$ were applied, with $A$ ranging from \qtyrange{10}{50}{\kilo\pascal} in \SI{10}{\kilo\pascal} increments and $\omega$ from \qtyrange[per-mode=symbol]{1}{6}{\radian\per\second} in increments of \SI[per-mode=symbol]{1}{\radian\per\second}, yielding 30 sine-input trials. The measured actuator pressure and robot position were used to evaluate the sensor placement framework, with the sensor configuration determined by the optimization problem.

One type of sensor considered were position sensors, which measured the deflection of the continuum arm. The seven centroids spaced \SI{2}{\centi\meter} apart, obtained from the arm's 2D projection, served as candidate sensor locations and formed the rows of the candidate $\mbf{C}$ matrix according to the measurement model in \eqref{eq:position_measurement_model}. The optimal sensor placement was determined using \eqref{eq:opt_prob_convex}, for $p=1,2,3$ active sensors, and all three objective functions listed in Section~\ref{sec:observability_tools}.

Strain was another type of measurement considered. The strain measurements were approximated from mocap data by fitting the beam's deflection $w(x)$ to the measured positions of the seven points along the arm. This provided a continuous strain function along the beam, and 13 candidate sensor locations were selected at \SI{1}{\centi\meter} intervals. As with position sensors, the optimal sensor placement was determined using all three objective functions, with $p=1,2,3$ active sensors.

To evaluate performance with the collected data, only measurements corresponding to the selected placement were used in a Kalman filter (KF) to estimate the robot's shape. Input pressure covariance was estimated by averaging the variance of the measured pressure signal across trials with constant pressures between \qtyrange[range-phrase={ and }]{15}{100}{\kilo\pascal}, in \SI{5}{\kilo\pascal} increments. For the position measurement's noise covariance, the mocap system's calibration error of \SI{0.5}{\milli\meter} was taken as the noise standard deviation. To account for marker misplacement relative to their modelled positions and errors from interpolating across occluded frames, this value was inflated by an order of magnitude when computing the measurement noise covariance. The noise covariance for the strain measurements, which were approximated from position data rather than directly measured, was based on the position sensor noise but inflated by two orders of magnitude to account for additional noise from the approximation and model uncertainty.

The ground truth was defined as the shape estimate obtained from the KF using all available mocap data, corresponding to the position measurements of all seven centroids shown in Fig.~\ref{fig:markers_and_centroids}.
This ground truth choice avoids relying directly on noisy mocap measurements affected by interpolation during occluded frames.
The estimated deflection profiles were evaluated at the marker locations and compared to the ground truth using RMSE.

The approach was experimentally evaluated using three test cases: one involving the optimal placement of position sensors (mocap markers), another focusing on strain sensors, and the third combining both sensor types. The first two cases evaluated performance for $p = 1, 2, 3$ active sensors under both step and sinusoidal inputs. In the third case, which involved placing 1 position sensor and 3 strain sensors, a constraint was added to exclude the last 2 of the 7 candidate position sensor locations and the last 4 of the 13 candidate strain sensor locations.
This constraint was added to emulate physical limitations that may arise when the end effector is used for manipulation tasks. For example, operating in a confined environment could cause occlusion at the end of the arm, preventing marker tracking, or the attachment of an end effector could limit the available space for sensor placement.
    \section{Results}
\label{sec:results}

\subsection{Position Sensor Placement}
\label{sec:pos_sensor_placement}
Let $\{ 1, 2, \ldots, 7 \}$ denote the indices of the candidate sensor locations, from the fixed base to the free end of the arm. Table~\ref{table:pos_opt_results} shows the optimal sensor placements for $p = 1, 2, 3$ and each objective function, along with a baseline placement of evenly spaced sensors.
In all cases, the objective function $\lambda_{\min}(\mbf{W}_{\mathrm{o}})$ resulted in a nonbinary selector vector, meaning no optimal configuration could be identified.
While the evenly spaced baseline is intuitive for multiple sensors, the tip location is intuitively optimal for a single position sensor due to its large deflection. However, the baseline placement was deliberately chosen as the middle location for a meaningful comparison. This case is included to show that the optimization recovers the expected optimal tip placement.

\begin{table}[!ht]
    \caption{Placement for $p$ position sensors and different $\mathcal{J}(\mbf{W}_{\mathrm{o}})$}
    \label{table:pos_opt_results}
    \centering
    \begin{tabularx}{\columnwidth}{
        X
        >{\centering\arraybackslash}X
        >{\centering\arraybackslash}X
        >{\centering\arraybackslash}X}
        \toprule
        $\mathcal{J}(\mbf{W}_{\mathrm{o}})$ & $p = 1$ & $p = 2$ & $p = 3$ \\
        \midrule
        $\lambda_{\min}(\mbf{W}_{\mathrm{o}})$ & N.A. & N.A. & N.A. \\
        $\trace(\mbf{W}_{\mathrm{o}})$ & $\{7 \}$ & $\{6, 7\}$ & $\{5, 6, 7\}$ \\
        $\log(\det(\mbf{W}_{\mathrm{o}}))$ & $\{7 \}$ & $\{4, 7\}$ & $\{4, 6, 7\}$ \\
        Baseline & $\{ 4 \}$ & $\{4, 7\}$ & $\{3, 5, 7\}$ \\
        \bottomrule
    \end{tabularx}
\end{table}

\vspace{-7pt}
\begin{table*}[!ht]
    \vspace{5pt}
    \begin{threeparttable}
        \caption{Total trajectory RMSE for position sensor placements from Table~\ref{table:pos_opt_results}, grouped by actuation input type}
        \label{table:rmse_traj_pos}
        \centering
        \begin{tabularx}{\textwidth}{X c c c c c c}
            \toprule
            \multirow{2}{*}{$\mathcal{J}(\mbf{W}_{\mathrm{o}})$} & \multicolumn{3}{c}{Step Input} & \multicolumn{3}{c}{Sinusoidal Input} \\
            \cmidrule(lr){2-4} \cmidrule(lr){5-7}
            & $p=1$ & $p=2$ & $p=3$ & $p=1$ & $p=2$ & $p=3$ \\
            \midrule
            $\trace(\mbf{W}_{\mathrm{o}})$ & \textbf{1.82}$^{***}$ (1.00, 3.27) & 1.05$^{*}$ (0.69, 1.97) & 0.62$^{***}$ (0.20, 1.17) & \textbf{1.34}$^{***}$ (0.86, 2.08) & 0.81$^{**}$ (0.57, 1.09) & 0.52 (0.32, 0.65) \\
            $\log(\det(\mbf{W}_{\mathrm{o}}))$ & \textbf{1.82}$^{***}$ (1.00, 3.27) & \textbf{0.97} (0.66, 1.55) & 0.59$^{***}$ (0.51, 1.04) & \textbf{1.34}$^{***}$ (0.86, 2.08) & \textbf{0.59} (0.39, 1.11) & \textbf{0.4} (0.29, 0.59) \\
            Baseline & 3.45 (1.23, 6.83) & \textbf{0.97} (0.66, 1.55) & \textbf{0.53} (0.14, 0.96) & 3.01 (2.04, 3.69) & \textbf{0.59} (0.39, 1.11) & 0.44 (0.27, 0.75) \\
            \bottomrule
        \end{tabularx}
        \begin{tablenotes}[flushleft]
            \item Values show median (Q1, Q3) RMSE in \SI{}{\milli\meter}. Wilcoxon signed-rank test significance vs. baseline ($p$-values): $^{*}<0.05$, $^{**}<0.01$, $^{***}<0.001$.
        \end{tablenotes}
    \end{threeparttable}
\end{table*}

\begin{figure}[ht]
    \centering
    \includegraphics[width=\linewidth]{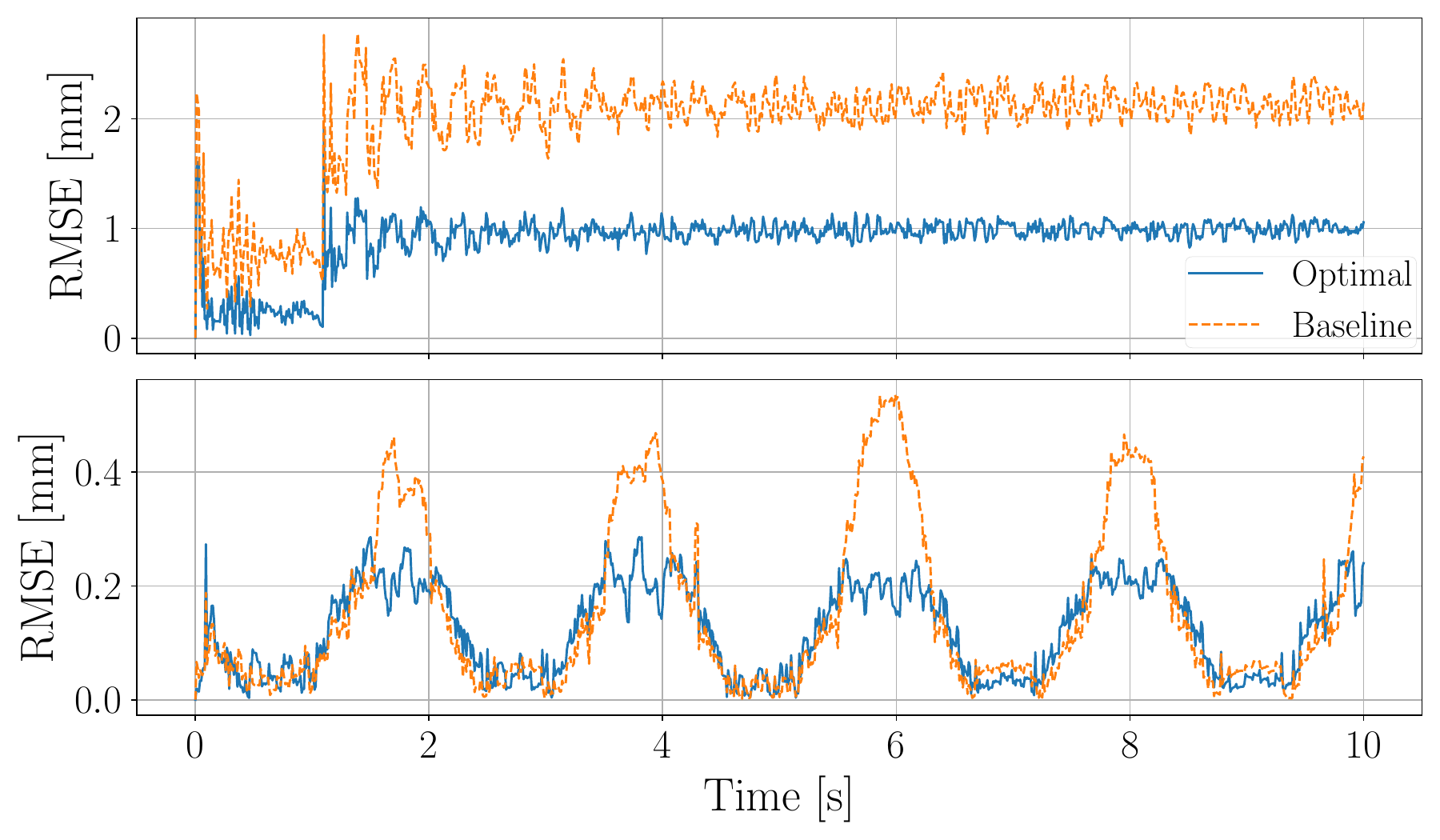}
    \vspace{-20pt}
    \caption{Time evolution of RMSE for representative trials from Section~\ref{sec:pos_sensor_placement}. Top: $p=1$, step input $p_{\text{set}}=$ \SI{90}{\kilo\pascal}. Bottom: $p=3$, sinusoidal input $p_{\text{set}}=40(\sin(3t) + 1)$ \SI{}{\kilo\pascal}, $\mathcal{J}(\mbf{W}_{\mathrm{o}})~=~\log(\det(\mbf{W}_{\mathrm{o}}))$.}
    \label{fig:representative_trials}
\end{figure}

\begin{figure*}[!ht]
    \centering
    \includegraphics[width=\textwidth]{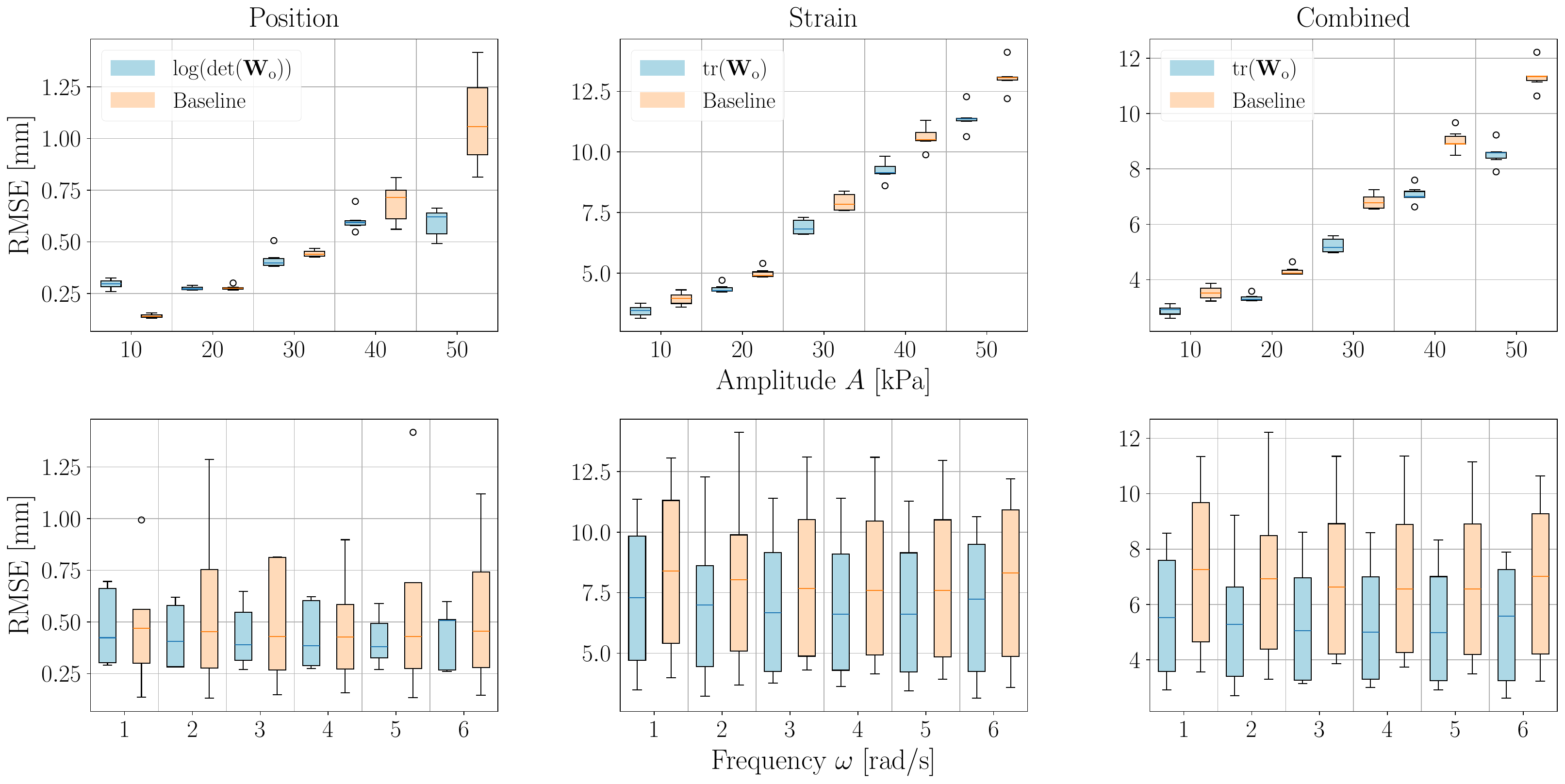}
    \vspace{-20pt}
    \caption{Total trajectory RMSE for different actuation amplitudes $A$ [\SI{}{\kilo\pascal}] and frequencies $\omega$ [\SI[per-mode=symbol]{}{\radian\per\second}] in $p_{\text{set}} = A(\sin(\omega t) + 1)$. Box plots show RMSE distributions across all sinusoidal trials. Left column: $p=3$ position measurements with $\log(\det(\mbf{W}_{\mathrm{o}}))$ placement vs. baseline. Middle column: $p=3$ strain measurements with $\trace(\mbf{W}_{\mathrm{o}})$ placement vs. baseline. Right column: $p_{1}=1$ and $p_{2}=3$ combined measurements with $\trace(\mbf{W}_{\mathrm{o}})$ placement vs. baseline.}
    \label{fig:rmse_traj_amp_freq_banner}
    \vspace{-7pt}
\end{figure*}

Using the placements in Table~\ref{table:pos_opt_results}, a KF was applied to reconstruct the arm's shape across all step and sinusoidal trials. The RMSE of the optimal placements was compared to the baseline. Representative RMSE plots are shown in Fig.~\ref{fig:representative_trials}, and the total trajectory RMSE across all trials is shown in Table~\ref{table:rmse_traj_pos}.
For step inputs, the $p=1$ tip placement yielded a lower RMSE than the baseline middle placement. However, for $p=2,3$, the optimized placement resulted in a higher RMSE than the baseline, except for the $p=2$ case with $\mathcal{J}(\mbf{W}_{\mathrm{o}})=\log(\det(\mbf{W}_{\mathrm{o}}))$, which matched the baseline.

The sinusoidal input results are particularly valuable for assessing sensor placement effectiveness since they provide a broader frequency excitation, unlike step inputs which are limited to low-frequency excitation. Under sinusoidal excitation, optimal placements outperformed the baseline for the $p=1$ and $p=3$ cases, while for $p=2$ the $\log(\det(\mbf{W}_{\mathrm{o}}))$ placement equaled baseline performance.

To explore how RMSE varied with actuation amplitude and frequency in greater detail, the left column of Fig.~\ref{fig:rmse_traj_amp_freq_banner} shows RMSE distributions across all sinusoidal input trials, grouped by these parameters. The analysis uses the $\log(\det(\mbf{W}_{\mathrm{o}}))$ placement since it was the best performing configuration for $p = 3$ under sinusoidal inputs. At higher amplitudes as well as higher frequencies, the optimal placement's RMSE was lower than the baseline's. Amplitude and frequency analysis box plots for all other sensor placement cases can be viewed in the source code repository.

\subsection{Strain Sensor Placement}
\label{sec:strain_sensor_placement}
With candidate sensor locations indexed $\{1, 2, \ldots, 13\}$ from base to tip, the optimal placements are listed in Table~\ref{table:strain_opt_results}, along with a baseline configuration using evenly spaced sensors. The same process as in Section~\ref{sec:pos_sensor_placement} was used to estimate the robot's shape, and the RMSE from the optimal placements was compared to the baseline.

\begin{table}[!ht]
    \caption{Sensor placement for $p$ strain sensors and different $\mathcal{J}(\mbf{W}_{\mathrm{o}})$}
    \label{table:strain_opt_results}
    \centering
    \begin{tabularx}{\columnwidth}{
        X
        >{\centering\arraybackslash}X
        >{\centering\arraybackslash}X
        >{\centering\arraybackslash}X}
        \toprule
        $\mathcal{J}(\mbf{W}_{\mathrm{o}})$ & $p = 1$ & $p = 2$ & $p = 3$ \\
        \midrule
        $\lambda_{\min}(\mbf{W}_{\mathrm{o}})$ & $\{1 \}$ & $\{1, 7\}$ & $\{1, 7, 8\}$ \\
        $\trace(\mbf{W}_{\mathrm{o}})$ & $\{1 \}$ & $\{1, 2\}$ & $\{1, 2, 3\}$ \\
        $\log(\det(\mbf{W}_{\mathrm{o}}))$ & $\{1 \}$ & $\{1, 2\}$ & $\{1, 2, 7\}$ \\
        Baseline & $\{ 7 \}$ & $\{1, 7\}$ & $\{1, 5, 9\}$ \\
        \bottomrule
    \end{tabularx}
    \vspace{-13pt}
\end{table}

Table~\ref{table:rmse_traj_strain} shows the total trajectory RMSE across all trials. For $p=1$, all three objective functions selected the base of the arm, yielding lower RMSE for both input types. For $p=2,3$, the $\trace(\mbf{W}_{\mathrm{o}})$ placements with concentrated sensors near the base, outperformed the baseline for both input types.

\begin{table*}[!ht]
    \vspace{5pt}
    \begin{threeparttable}
        \caption{Total trajectory RMSE for strain sensor placements from Table~\ref{table:strain_opt_results}, grouped by actuation input type}
        \label{table:rmse_traj_strain}
        \centering
        \setlength{\tabcolsep}{1.5pt}
        \begin{tabularx}{\textwidth}{X c c c c c c}
            \toprule
            \multirow{2}{*}{$\mathcal{J}(\mbf{W}_{\mathrm{o}})$} & \multicolumn{3}{c}{Step Input} & \multicolumn{3}{c}{Sinusoidal Input} \\
            \cmidrule(lr){2-4} \cmidrule(lr){5-7}
            & $p=1$ & $p=2$ & $p=3$ & $p=1$ & $p=2$ & $p=3$ \\
            \midrule
            $\lambda_{\min}(\mbf{W}_{\mathrm{o}})$ & \textbf{11.46}$^{***}$ (5.24, 20.98) & 11.55 (5.28, 21.15) & 12.12$^{***}$ (5.54, 22.20) & \textbf{8.82}$^{***}$ (5.50, 12.15) & 8.88 (5.52, 12.23) & 9.31$^{***}$ (5.78, 12.83) \\
            $\trace(\mbf{W}_{\mathrm{o}})$ & \textbf{11.46}$^{***}$ (5.24, 20.98) & \textbf{10.14}$^{***}$ (4.63, 18.56) & \textbf{8.88}$^{***}$ (4.06, 16.26) & \textbf{8.82}$^{***}$ (5.50, 12.15) & \textbf{7.80}$^{***}$ (4.86, 10.74) & \textbf{6.83}$^{***}$ (4.25, 9.41) \\
            $\log(\det(\mbf{W}_{\mathrm{o}}))$ & \textbf{11.46}$^{***}$ (5.24, 20.98) & \textbf{10.14}$^{***}$ (4.63, 18.56) & 10.26$^{***}$ (4.68, 18.78) & \textbf{8.82}$^{***}$ (5.50, 12.15) & \textbf{7.80}$^{***}$ (4.86, 10.74) & 7.88$^{***}$ (4.9, 10.86) \\
            Baseline & 17.65 (14.17, 34.78) & 11.55 (5.28, 21.15) & 10.21 (4.66, 18.69) & 15.75 (13.18, 19.13) & 8.88 (5.52, 12.23) & 7.84 (4.87, 10.81) \\
            \bottomrule
        \end{tabularx}
        \begin{tablenotes}[flushleft]
            \item Values show median (Q1, Q3) RMSE in \SI{}{\milli\meter}. Wilcoxon signed-rank test significance vs. baseline ($p$-values): $^{*}<0.05$, $^{**}<0.01$, $^{***}<0.001$.
        \end{tablenotes}
    \end{threeparttable}
    \vspace{-10pt}
\end{table*}

Across input types, the relative ranking of placements is consistent, with the same configurations achieving the lowest RMSE under both step and sinusoidal inputs. As in the position sensor case, step inputs mainly excite low-frequency dynamics, whereas sinusoidal inputs excite a broader range and are therefore more informative for evaluating placement. Accordingly, the sinusoidal input trials are analyzed in greater detail to assess how actuation amplitude and frequency influence the total trajectory RMSE. This analysis uses the $\trace(\mbf{W}_{\mathrm{o}})$ optimal placement, which had the lowest RMSE for $p=3$ active strain sensors under sinusoidal inputs.
As shown in the middle column of Fig.~\ref{fig:rmse_traj_amp_freq_banner}, the optimal placement's better performance becomes more pronounced at higher amplitudes, and yields consistently lower RMSE than the baseline across all tested frequencies.

\subsection{Combined Sensor Placement}
\label{sec:combined_sensor_placement}
This test case used both position and strain sensors with the optimization problem~\eqref{eq:opt_prob_convex_2_types}. The optimal placements for $p_{1} = 1$ position sensor and $p_{2} = 3$ strain sensors are listed in Table~\ref{table:pos_strain_opt_results}. The objective function $\lambda_{\min}(\mbf{W}_{\mathrm{o}})$ yielded a nonbinary selector vector for the position sensor, meaning no optimal placement was found. The baseline placement again used evenly spaced sensors: one position sensor at the middle of the arm and three strain sensors evenly distributed.

\begin{table}[!ht]
    \caption{Resulting placement for $p_{1}$ position sensors and $p_{2}$ strain sensors, using different $\mathcal{J}(\mbf{W}_{\mathrm{o}})$}
    \vspace{-5pt}
    \label{table:pos_strain_opt_results}
    \centering
    \begin{tabularx}{0.8\columnwidth}{
        X
        >{\centering\arraybackslash}X
        >{\centering\arraybackslash}X}
        \toprule
        $\mathcal{J}(\mbf{W}_{\mathrm{o}})$ & $p_{1}=1$ & $p_{2}=3$ \\
        \midrule
        $\lambda_{\min}(\mbf{W}_{\mathrm{o}})$ & N.A. & $\{1, 7, 8 \}$ \\
        $\trace(\mbf{W}_{\mathrm{o}})$ & $\{5 \}$ & $\{1, 2, 3\}$ \\
        $\log(\det(\mbf{W}_{\mathrm{o}}))$ & $\{5 \}$ & $\{1, 2, 7\}$ \\
        Baseline & $\{ 4 \}$ & $\{1, 5, 9\}$ \\
        \bottomrule
    \end{tabularx}
    \vspace{-5pt}
\end{table}

Following the same process as in Sections~\ref{sec:pos_sensor_placement}~and~\ref{sec:strain_sensor_placement}, the robot's shape was estimated across all trials, and the RMSE from the optimal placements was compared to the baseline. As shown in Table~\ref{table:rmse_traj_pos_strain}, both the $\trace(\mbf{W}_{\mathrm{o}})$ and $\log(\det(\mbf{W}_{\mathrm{o}}))$ placements achieved a lower RMSE across all trials, with the $\trace(\mbf{W}_{\mathrm{o}})$ placement performing the best.

The right column of Fig.~\ref{fig:rmse_traj_amp_freq_banner} shows the results of the amplitude and frequency analysis applied to the $\trace(\mbf{W}_{\mathrm{o}})$ sensor placement. Consistent with the previous sections, the optimal placement yields lower RMSE than the baseline at higher amplitudes and maintains a consistent advantage across the tested frequencies.

\begin{table}[!ht]
    \begin{threeparttable}
        \caption{Total trajectory RMSE for combined sensor placement from Table~\ref{table:pos_strain_opt_results}, grouped by actuation input type.}
        \label{table:rmse_traj_pos_strain}
        \centering
        \begin{tabularx}{\columnwidth}{X c c}
            \toprule
            $\mathcal{J}(\mbf{W}_{\mathrm{o}})$ & Step Input & Sinusoidal Input\\
            \midrule
            $\trace(\mbf{W}_{\mathrm{o}})$ & \textbf{6.81}$^{***}$ (3.55, 12.47) & \textbf{5.16}$^{***}$ (3.25, 7.19)\\
            $\log(\det(\mbf{W}_{\mathrm{o}}))$ & 7.49$^{***}$ (3.94, 13.72) & 5.67$^{***}$ (3.56, 7.9) \\
            Baseline & 8.7 (3.84, 16.06) & 6.78 (4.21, 9.18) \\
            \bottomrule
        \end{tabularx}
        \begin{tablenotes}[flushleft]
            \item Values show median (Q1, Q3) RMSE in \SI{}{\milli\meter}. Wilcoxon signed-rank test significance vs. baseline ($p$-values): $^{*}<0.05$, $^{**}<0.01$, $^{***}<0.001$.
        \end{tablenotes}
    \end{threeparttable}
    \vspace{-10pt}
\end{table}
\FloatBarrier
    \section{Discussion}
\label{sec:discussion}
Placing a single position sensor at the tip consistently yielded lower RMSE than the middle placement, likely because the tip undergoes the largest deflections and provides more informative measurements. The higher RMSE observed with the $\trace(\mbf{W}_{\mathrm{o}})$ placement for $p = 2$ and $p = 3$ may be due to the fact that maximizing $\trace(\mbf{W}_{\mathrm{o}})$ does not prevent weakly observable modes. The minimum eigenvalue of $\mbf{W}_{\mathrm{o}}$ was lower than that of the baseline in these cases, suggesting poor observability of the second mode. Meanwhile, the $\log(\det(\mbf{W}_{\mathrm{o}}))$ placement performed slightly worse than the baseline for $p = 3$ under step inputs, but outperformed it under sinusoidal inputs.
Since the robot is modelled as a linear beam, the model more accurately captures small deflections than large ones.
As shown in Fig.~\ref{fig:rmse_traj_amp_freq_banner}, the $\log(\det(\mbf{W}_{\mathrm{o}}))$ position sensor placement better mitigates this modelling error at higher actuation amplitudes and frequencies, indicating that increased observability may add robustness to model error.

Placing a single strain sensor at the base consistently yielded lower RMSE than the baseline middle placement for both types of inputs, consistent with the fact that the highest strain in a cantilever beam occurs near the base. For the $p=2,3$ cases with the $\trace(\mbf{W}_{\mathrm{o}})$ placement, the results suggest that concentrating multiple sensors in the region modelled to experience the most strain is more informative than placing them across separate locations.
This finding is supported by the high statistical significance of the results in Table~\ref{table:rmse_traj_strain}, where all $\trace(\mbf{W}_{\mathrm{o}})$ placements outperformed the baseline with $p$-values~$<$~0.001.
As in the position sensor case, the placement with maximal observability is better at handling modelling error at higher actuation amplitudes.

An advantage of the optimization problem~\eqref{eq:opt_prob_convex} is that it allows for built-in physical constraints on sensor placement. This was demonstrated in the combined sensing experiment, where certain locations were excluded from the candidate set to emulate physical or spatial limitations. These constraints were easily handled by removing those locations from the set used to construct the initial $\mbf{C}$ matrix. In this test case, the $\trace(\mbf{W}_{\mathrm{o}})$ and $\log(\det(\mbf{W}_{\mathrm{o}}))$ placements outperformed the baseline, which may be attributed to the stronger contribution of position sensors to the overall observability of the system.

The optimal placement in the $p=1$ position sensor case is intuitive; it has already been shown that the full robot state can be recovered from tip velocities and a nonlinear observer~\cite{zheng2024full}. The proposed method is particularly beneficial when placing multiple sensors or sensor types, as the placement in these cases is less intuitive.

The sinusoidal input analyses indicate that placements selected using linear observability tools still provide better results even under increased nonlinearity at higher actuation amplitudes. In practice, smooth actuation is typically preferred and step inputs mainly excite low-frequency dynamics, so performance under sinusoidal inputs is a more informative benchmark. Moreover, sinusoids form a basis for periodic signals via Fourier series, so strong sinusoidal performance suggests broader applicability to periodic actuation.

No single observability metric consistently outperforms the others. All metrics recover the optimal placement for the $p=1$ cases, but for $p=2,3$ the best placement alternates between the $\trace(\mbf{W}_{\mathrm{o}})$ and $\log(\det(\mbf{W}_{\mathrm{o}}))$ placements depending on the sensor type. Metric choice is not universal and may depend on sensor count and type.

In principle, this sensor placement method is generalizable to soft robots that can be modelled as linear state-space systems. The optimization problem requires only the $\mbf{A}$ and $\mbf{C}$ matrices to determine the optimal sensor placement. The main challenge in applying this method to other soft robot morphologies is identifying an appropriate state representation and dynamic model. If necessary, the model must then be manipulated and linearized to obtain a linear state-space system. In practice, a simplified linear model may suffice for sensor placement, even if it lacks the accuracy required for state estimation or control. A more accurate nonlinear model could then leverage these optimally placed sensors for such tasks. There is precedent for the linearization of soft robots, such as the finite element model of a soft trunk linearized to enable gain-scheduled control \cite{wuSoftManipulatorControl2021}.

Despite the method's generalizability, several practical limitations remain. The selector vector resulting from the relaxed optimization problem must be binary, which is not always guaranteed. Furthermore, the method requires a state-space model in which at least one placement yields a positive-definite observability Gramian, which may be difficult to obtain for some soft robots. Future work will involve testing the method on additional robot morphologies and sensing modalities.

    \section{Conclusion}
\label{sec:conclusion}
This paper implements and experimentally evaluates a generalizable sensor placement framework that accounts for system dynamics, using a continuum arm as a representative example of a soft robot.
The method uses convex optimization to maximize system observability using different Gramian-based observability metrics. 
In most cases, the resulting optimal placements either matched or outperformed the baseline configuration in terms of RMSE when estimating the robot's shape.
Notably, performance remains strong under larger deflections, where the linear model would typically fail, indicating potential robustness to modelling errors.
Even when performance matched the baseline, the method demonstrated its value by automating sensor placement. This feature is particularly useful for soft robots with many potential sensor locations, where intuitive placement is unclear.
The results further demonstrate that linear observability tools can be applied to this soft continuum arm with strong performance, suggesting an effective approach that may extend to other inherently nonlinear soft robots.

    \printbibliography
\end{document}